\documentclass[letterpaper]{article} 
\usepackage{aaai25}  
\usepackage{times}  
\usepackage{helvet}  
\usepackage{courier}  
\usepackage[hyphens]{url}  
\usepackage{graphicx} 
\usepackage{natbib}  
\usepackage{caption} 
\usepackage{algorithm}
\usepackage{algorithmic}

\usepackage{newfloat}
\usepackage{listings}
\DeclareCaptionStyle{ruled}{labelfont=normalfont,labelsep=colon,strut=off} 
\floatstyle{ruled}
\newfloat{listing}{tb}{lst}{}
\floatname{listing}{Listing}

\usepackage{amssymb}
\usepackage{booktabs}
\usepackage{multicol}
\usepackage{multirow}
\usepackage[table]{xcolor}         
\usepackage{amsmath}
\usepackage{enumitem}
\definecolor{Gray}{gray}{0.9}

\title{Densely Connected Parameter-Efficient Tuning for Referring Image Segmentation}
\author{
    Jiaqi Huang$^*$\textsuperscript{\rm 1,2},
    Zunnan Xu$^*$\textsuperscript{\rm 1,2},
    Ting Liu$^\dag$\textsuperscript{\rm 1},
    Yong Liu\textsuperscript{\rm 1},
    Haonan Han\textsuperscript{\rm 1},
    Kehong Yuan$^{\ddag}$\textsuperscript{\rm 1},
    Xiu Li\textsuperscript{\rm 1},
}

\affiliations{
    \textsuperscript{\rm 1}Tsinghua Shenzhen International Graduate School, Tsinghua University, \textsuperscript{\rm 2}Tencent\\

    University Town of Shenzhen, Nanshan District, 
    Shenzhen, Guangdong, P.R. China\\
    \{huangjq23,xzn23\}@mails.tsinghua.edu.cn, yuankh@sz.tsinghua.edu.cn
}

\begin{document}

\maketitle

\def\thefootnote{*}\footnotetext{Equal Contribution. Work done as interns at Tencent.}
\def\thefootnote{\dag}\footnotetext{Work done during Ting Liu's visit at Tsinghua University.}
\def\thefootnote{\ddag}\footnotetext{Corresponding author}

\begin{abstract}
In the domain of computer vision, Parameter-Efficient Tuning (PET) is increasingly replacing the traditional paradigm of pre-training followed by full fine-tuning. PET is particularly favored for its effectiveness in large foundation models, as it streamlines transfer learning costs and optimizes hardware utilization.
However, the current PET methods are mainly designed for single-modal optimization.
While some pioneering studies have undertaken preliminary explorations, they still remain at the level of aligned encoders (e.g., CLIP) and lack exploration of misaligned encoders.
These methods show sub-optimal performance with misaligned encoders, as they fail to effectively align the multimodal features during fine-tuning.
In this paper, we introduce DETRIS, a parameter-efficient tuning framework designed to enhance low-rank visual feature propagation by establishing dense interconnections between each layer and all preceding layers, which enables effective cross-modal feature interaction and adaptation to misaligned encoders. We also suggest using text adapters to improve textual features. Our simple yet efficient approach greatly surpasses state-of-the-art methods with 0.9\% to 1.8\% backbone parameter updates, evaluated on challenging benchmarks. Our project is available at \url{https://github.com/jiaqihuang01/DETRIS}.

\end{abstract}

\section{Introduction} \label{introduction}

Referring Image Segmentation (RIS) aims to predict the mask of a target object within an image based on a natural language description.
Unlike semantic segmentation, which involves assigning a label from a predefined set to each pixel in an image, RIS requires a more nuanced understanding of the language and visual content to identify the described object.
The RIS task holds great significance as it effectively bridges the gap between natural language descriptions and fine-grained visual perception~\cite{ji2024survey}.
This capability is crucial for advancing the field of artificial intelligence, particularly in areas such as autonomous systems, image-based retrieval, and human-computer interaction.
The complexity of RIS arises from the need to interpret arbitrary context lengths and to comprehend an open-world vocabulary that includes a wide array of object names, attributes, and positional references~\cite{li2024text}.
The requirement for precise segmentation of referring objects makes this dense prediction task one of the most challenging tasks in vision language understanding.

In the field of Computer Vision, scaling up foundational models~\cite{radford2021learning,li2022grounded,ma2022simvtp,oquab2023dinov2,fang2023eva} is becoming increasingly important. These models leverage large datasets to learn a comprehensive set of visual features. The scaling process not only enhances their ability to discern subtleties in visual data but also significantly boosts their generalization capabilities. With more parameters and exposure to a wider range of data, these models are better equipped to handle diverse and complex visual tasks~\cite{ma2022visual,he2023camouflaged,he2024weakly,fang2024real,zhuang2025kdpror}, demonstrating robustness that is essential for real-world applications~\cite{he2024diffusion}.

However, there often exists a gap between the pre-trained tasks of these models and the specific requirements of downstream applications. Bridging this gap through efficient adaptation presents a formidable challenge.
Recent studies~\cite{cris,ding2022vlt,lavt,gres,li2023g2l} have demonstrated the effectiveness of fine-tuning powerful pre-trained models for referring image segmentation.
However, a common challenge is that they typically require full fine-tuning to adapt to dense prediction tasks. This process can lead to the loss of valuable pretraining knowledge, as it involves adjusting a large number of parameters that were previously optimized during the pre-training phase~\cite{restr,gres,polyformer,wu2024uncertainty}.
Moreover, these approaches maintain a distinct set of fine-tuned parameters of pretrained models for each dataset, which can lead to substantial deployment costs.
The problem becomes particularly serious when considering the ever-growing size of pre-trained models, which now include parameters ranging from hundreds of millions to trillions~\cite{li2022blip,zhou2022learning,chen2022pali,EVA-CLIP-18B}.

Various parameter-efficient tuning methods have been developed to achieve an optimal balance between operational efficiency and model performance~\cite{gao2021clip,chen2022adaptformer,zhou2022learning,wang2023barleria,li2024exploiting,liu2024dara}.
However, despite these contributions, most existing methods are limited in their scope, predominantly applied to single-modality tasks~\cite{guo2020parameter,houlsby2019parameter,chen2022vision} or simple classification problems~\cite{gao2021clip,chen2022adaptformer,zhou2022learning}. 
There remains a notable gap in research concerning dense prediction tasks and the nuanced interactions between multiple modalities. 
Pioneering works such as ETRIS and BarleRIa~\cite{wang2023barleria} aimed to parameter-efficient fine-tuning CLIP~\cite{radford2021learning} for referring image segmentation, but they encountered several limitations: 
(i) These methods primarily relied on the early-stage fusion of multimodal features from the backbone, missing out on the benefits of more comprehensive global features, which led to suboptimal results. 
(ii) 
Furthermore, existing parameter-efficient modules, such as Bridger~\cite{xu2023bridging} and GST~\cite{wang2023barleria}, are constrained by their limited application of multi-scale modeling. 
These approach directly fuses information from both modalities, which is insufficient for capturing the full complexity of visual data across different scales.

Our method addresses this question by introducing a simple yet efficient approach that enhances the effectiveness of adapting pre-trained vision-language models through multi-scale modeling and the incorporation of global prior information.
In detail, we propose an adapter named Dense Aligner, which can be seamlessly integrated into pre-trained models for dense prediction tasks.
There are two customized modules for Dense Aligner:
(i) a dense mixture of convolution module designed to capture multi-scale semantic features from the intermediate layers and (ii) a cross-aligner module that facilitates information exchange between visual and textual features.
Secondly, we propose incorporating text adapters to enhance the text encoder. We further leverage these enhanced features to improve alignment between visual and linguistic features.

Our framework is built around a dual encoder architecture. Unlike previous PEFT methods that rely on highly aligned encoders (e.g., CLIP), we support DINO~\cite{oquab2023dinov2} as our visual encoder.
The reason we chose DINO as our vision backbone is based on several insights:
(i) DINO's self-supervised learning approach provides robust generalization and is especially beneficial for dense prediction tasks compared to CLIP~\cite{radford2021learning}.
(ii) The absence of multimodal pre-training in DINO, particularly in visual-text alignment, presents challenges for its direct application on referring image segmentation. This deficiency underscores the essential role of our proposed module in boosting the model's abilities, especially in enhancing visual language alignment and the execution of dense prediction tasks.

Our main contributions are as follows:
\begin{itemize}[leftmargin=*,noitemsep,nolistsep]
    \item We support the powerful pre-trained model DINO in RIS tasks and provide a parameter-efficient fine-tuning strategy for visual-text alignment that avoids the need for intricate design.
    \item We propose a simple yet efficient adapter called Dense Aligner that can be seamlessly integrated into the pre-trained backbone to enhance and interact with its intermediate features. This integration improves DINO's alignment with language and enhances its performance on dense prediction tasks.
    \item Experiments demonstrate that our method greatly surpasses state-of-the-art fully fine-tuned methods in referring image segmentation, with only 0.9\% to 1.8\% backbone parameter updates.
\end{itemize}

\begin{figure*}[htbp]
    \centering
    \includegraphics[width=1\linewidth]{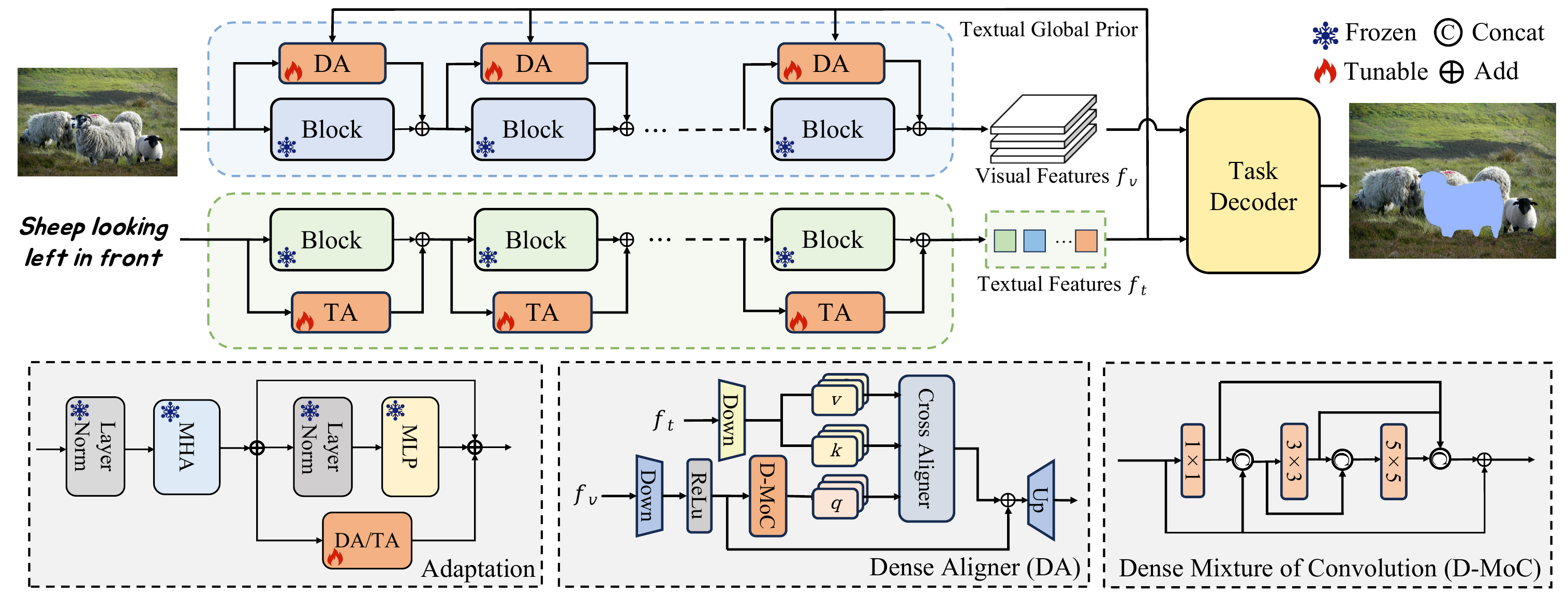}
    \caption{Overall framework of our DETRIS. In the image branch, we utilize Dense Aligner (DA) to facilitate cross-modal and multi-scale modeling of low-rank visual features. This approach incorporates textual global prior information to enhance the visual features~\( f_v \). In the text branch, we also use ``D-MoC'' as our Text Adapters (TA) to obtain the text feature \( f_t \).}
    \label{fig:model}
\end{figure*}

\section{Related Work}

\textbf{Parameter Efficient Tuning (PET)} aims to streamline the process of adapting pre-trained models to new tasks with minimal parameter adjustments, making it a practical solution for deploying large models to individual users, particularly in the face of expanding model sizes. 
Previous PET methods can be mainly divided into three types:
(i) updating newly added parameters to the model or input~\cite{houlsby2019parameter,li2021prefix,zhou2022learning};
(ii) sparsely updating a small number of parameters of the model~\cite{guo2020parameter,zaken2021bitfit};
(iii) low-rank factorization for the weights to be updated~\cite{hu2021lora,karimi2021compacter,haoconsolidator}. 
However, previous works applying PET in computer vision mainly focus on classification and generation tasks.
How to efficiently update and transfer the pre-trained knowledge space to dense prediction tasks remains a great challenge.
Some pioneering work like ETRIS~\cite{xu2023bridging} and BarleRIa~\cite{wang2023barleria} sought to utilize adapters to fine-tune CLIP~\cite{radford2021learning} for referring image segmentation. However, their proposed modules like Bridger~\cite{xu2023bridging} and GST~\cite{wang2023barleria} are insufficient for capturing the complexity of multi-scale visual features.

\textbf{Referring Image Segmentation (RIS)}
aims to segment the target objects referred to by natural language descriptions.
It necessitates the models to comprehensively associate diverse visual content and linguistic signals.
The genesis of this field can be traced to CNN-LSTM-based methods, such as the Referring Relation Network (RRN)~\cite{li2018referring} and the Recurrent Multimodal Interaction (RMI)~\cite{liu2017recurrent}. These early methods used CNN and LSTM networks to separately extract visual and linguistic features, which were then combined to form cross-modal representations for segmentation by an FCN.
The advent of the Transformer model has catalyzed a paradigm transformation of integrating features across diverse modalities by attention mechanism~\cite{lavt, gres, uninext, polyformer}.
Among them, MDETR~\cite{kamath2021mdetr} and VLT~\cite{ding2022vlt} have demonstrated remarkable performance across various Vision-Language (VL) tasks by integrating multi-modal attention interaction and query representation.
Capitalizing on the robust image-text alignment capabilities of CLIP, CRIS~\cite{cris}, ETRIS~\cite{xu2023bridging} and UniLSeg~\cite{liu2023universal} zeroes in on sentence-pixel alignment to harness the wealth of multi-modal correspondences. 
However, existing methods primarily concentrate on the design of visual-linguistic interactions during the decoding phase, while overlooking the potential to fully excavate the pretrained backbone networks.
To this end, we propose DETRIS, a framework that aligns features from different modalities with the assistance of parameter-efficient modules boosting multi-scale comprehensive updating of backbone networks. 
Compared to existing fully fine-tuning methods, the proposed approach achieves competitive performance while greatly reducing the training computation overhead.

\section{Methodology}
\subsection{Framework Overview}
The overall framework of the proposed model is illustrated in Figure~\ref{fig:model}. 
Our approach freezes the parameters of the pre-trained backbone, ensuring parameter efficiency. The core of the model's design philosophy is the Dense Aligner, which is intended to facilitate interaction between cross-modal features and inject dense prediction priors into the pre-trained backbone. This is coupled with the incorporation of text adapters in the text encoder to enhance fine-grained image-text alignments.

\subsection{Image \& Text Feature Extraction}
\textbf{Visual encoder.} 
Our work adapts the distilled DINOv2 with registers~\cite{oquab2023dinov2} as the backbone. This model has been well-pretrained on self-supervised training tasks and is based on ViT-B/14.
Specifically, for an input image $I \in R^{H\times W\times3}$, we utilize DINOv2 enhanced with Dense Aligners to extract image features. We select the outputs of several middle layers and the last layer as hierarchical vision features $f_\upsilon^i, i\in[1,2,3]$, which are used for subsequent dense prediction modules.

\textbf{Text encoder.} 
For the input refer expression $T$, we utilize the pre-trained text transformer of CLIP~\cite{radford2021learning} to extract text features.
Text features $f_{t}$ and sentence-level feature \( f_s \) are extracted using a CLIP text encoder, augmented with text adapters to guide referring image segmentation.

Considering the substantial number of parameters that the encoder occupies, and to avoid the loss of valuable pre-training knowledge, we freeze all the encoder parameters during our fine-tuning process to efficiently apply it in downstream tasks.

\subsection{Local \& Global Feature Interaction}
As mentioned in Section~\ref{introduction}, although DINOv2 has strong generalization capabilities and is more advantageous than CLIP in tasks that rely more on visual abilities, DINOv2 lacks visual-text alignment in the RIS downstream task due to the absence of multimodal pre-training. 
To address this and enhance the model's multi-scale modeling capability, we designed and utilized Dense Aligners to augment the model's vision backbone. 
The visual backbone network remains fixed, with training solely focused on the Dense Aligner parameters.

\textbf{Dense Aligner.} 
As shown in Figure~\ref{fig:model}, the proposed Dense Aligner differs significantly from previous adapter designs by incorporating a dense mixture of convolution modules. Additionally, a cross-aligner module is integrated between the activation and up-projection layers. This integration enhances the model's ability to extract dense image features and enriches its multimodal fusion capabilities.

In the dense mixture of convolution module, a multi-branch convolutional structure is introduced to effectively model low-rank visual features across multiple scales and to capture multi-scale prior information. The module sequentially applies linear projection and non-linear activation, followed by 1$\times$1, 3$\times$3, and 5$\times$5 convolutional kernels to progressively integrate outputs from previous layers. To maintain efficiency and reduce computational load, 1$\times$1 convolutions are placed before the 3$\times$3 and 5$\times$5 convolutions to compress the channel dimensions. 

In this dense mixture module, outputs from smaller kernels serve as inputs for larger kernels, ensuring a detailed stepwise integration of fine and broad features. The 1$\times$1 convolution output feeds into the 3$\times$3 convolution, and the combined outputs of the 1$\times$1 and 3$\times$3 convolutions are then passed to the 5$\times$5 convolution. This method differs from conventional direct merging of multi-scale features by delicately combining fine details with broader contextual information. To preserve the original features, the initial inputs are added back to the final concatenated outputs.

Specifically, for given input image features \(f_v^l\) at layer $l$, this process can be formulated  as below:
\begin{equation}
\begin{aligned}
&F_v^l = \sigma(\text{Linear}_\text{down}(f_v^l)), \\
&F_{v1}^l, F_{v2}^l, F_{v3}^l = \text{D-MoC}(F_v^l) \\
&F_{\text{dense}}^l = (F_{v1}^l, F_{v2}^l, F_{v3}^l) + F_v^l,
\end{aligned}
\end{equation}
where \(\sigma\) denotes non-linear activation function ReLU, \(\text{Linear}_\text{down}\) denotes a downsampling operation of linear projection. \(F_{\text{dense}}^l\) is the final dense feature representation obtained by concatenating the features from all branches and adding the initial feature \(F_v^l\). Here, \((,)\) represents concatenating the features along the dimensional direction.

For the $\text{D-MoC}$ operation, the process can be formulated as below:
\begin{equation}
\begin{aligned}
&F_{v1}^l = \text{conv}_{1 \times 1}(F_v^l), \\
&F_{v2}^l = \text{conv}_{3 \times 3}(F_v^l, F_{v1}^l), \\
&F_{v3}^l = \text{conv}_{5 \times 5}(F_v^l, F_{v1}^l, F_{v2}^l), \\
\end{aligned}
\end{equation}
where \(F_{v1}^l\), \(F_{v2}^l\), and \(F_{v3}^l\) are the output features after applying respective convolutional operations.

Considering that textual information contains valuable references, we utilize it as a global reference prior by integrating it into the vision backbone network via a aligner module. This not only regularizes the visual features but also aligns them better with the global features of the text (denoted as \( f_t \)). 
This process can be formalized as:
\begin{equation}
\begin{aligned}
&F_{\text{cross}}^l = \mathcal{F}_{\text{align}}(F_{\text{dense}}^l, f_t) + F_v^l, \\
&f_{dc}^l = \text{Linear}_\text{up}(F_{\text{cross}}^l),
\end{aligned}
\end{equation}
where \(\mathcal{F}_{\text{align}}\) represents the alignment method, and we found that simple multi-head cross attention is quite effective, 
\(F_{\text{cross}}^l\) denotes the fused visual feature, and \(\text{Linear}_\text{up}\) represents an operation to project the visual features back to get \(f_{dc}^l\).
We add the Dense Aligner in parallel to the MLP layer in the transformer block to achieve our adaptation as illustrated in Figure\ref{fig:model}.

\textbf{Text Adapter.} 
Acknowledging the disparities between the text features extracted by the CLIP text encoder and the DINO features, we incorporate a text adapter to improve the text encoder for fine-grained alignment of text and visual features.
In a manner similar to the Dense Aligner, we employ a Dense mixture of Convolution as our text adapter to better capture multi-scale text information and enhance the alignment between visual and textual features. However, unlike the Dense Aligner, which uses 2D convolutions, our text adapter leverages 1D convolutions specifically designed for processing textual data. This approach ensures that text features are effectively captured and integrated, enabling improved alignment and compatibility with the overall model architecture while optimizing the representation of sequential textual information.
Specifically, for given input text features \(f_t^l\) at layer $l$, this process can be formalized as:
\begin{equation}
\begin{aligned}
&F_t^l = \text{Linear}_\text{down}(f_t^l), \\
&F_{\text{relu}}^l = \text{D-MoC}(F_t^l), \\
&f_{\text{w}}^l = \text{Linear}_\text{up}(F_{\text{relu}}^l),
\end{aligned}
\end{equation}
where \(\text{Linear}_\text{down}\) represents a downsampling linear projection and \(\text{Linear}_\text{up}\) denotes an upsampling operation to adapt text features back to the original dimension.

\subsection{The Referring Image Segmentation Head}
To ensure a fair comparison, we follow CRIS~\cite{cris} and ETRIS~\cite{xu2023bridging}, incorporating a learnable referring image head. This head which consists of three main components: a cross-modal neck, a vision-language decoder, and an up-sample projector. These collaborate together to extract the cross-modal feature $F_c$ and the transformed textual feature $F_l$.

The cross-modal neck takes multiple adapted visual features ($\hat{f}_\upsilon^i, i\in[1,2,3]$) from three layers of the visual encoder (e.g., the 1/3, 2/3, and the last layer of the backbone) and the adapted textual embeddings $\hat{f}_t$. 
Specifically, we employ a multi-head cross-attention mechanism ($\mathcal{F}_{\text{MHCA}}$) with convolution to fuse these features, obtaining the fusion features $F_f$. Subsequently, we concatenate a 2D spatial coordinate feature $F_{\text{coord}}$ with $F_f$ and further fuse them using a 3$\times$3 convolution, which can be formalized as:
\begin{equation}
    f_c = \text{Conv}([F_f, F_{\text{coord}}]),
\end{equation}
where \(F_f = \mathcal{F}_{\text{MHCA}}(\hat{f}_v^i, \hat{f}_t)\) and \(f_c\) denotes the combined cross-modal feature.

The vision-language decoder further merges the composite feature \( f_c \) with the textual embeddings $\hat{f}_t$. This fusion process culminates in the generation of multimodal features \( F_{mm} \), encapsulating both visual and linguistic information.
Specifically, the decoder consists of three layers, each composed of a multi-head self-attention layer (MHSA), a multi-head cross-attention layer (MHCA), and a feed-forward network. 
Within each decoder layer, the combined features \( f_c \) are fed into the MHSA layer to capture global contextual information. The MHCA layer further facilitates multi-modal interaction by mapping visual features to queries and textual features to keys and values. 
Following the MHCA layer, an MLP block, along with layer normalization and residual connections, further processes the output features.

\begin{table*}[ht]
\caption{State-of-the-art comparison of RIS methods and the PET RIS method on RefCOCO/RefCOCO+/G-Ref datasets without using extra data and Mixed RefCOCO dataset, evaluated using the IoU metric. 
For Mixed RefCOCO datasets, models marked with \textsuperscript{*} are tuned using the mixed RefCOCO/RefCOCO+/G-Ref datasets.
The best results are in bold.}
\label{tab:main}
\centering
\setlength{\tabcolsep}{1.3mm} 
\begin{tabular}{l|cccccccccc}
\toprule
\multirow{2}{*}{Method} & \multicolumn{3}{c|}{RefCOCO} & \multicolumn{3}{c|}{RefCOCO+} & \multicolumn{3}{c|}{G-Ref} & \multirow{2}{*}{Avg} \\ 
\cline{2-10}
& \multicolumn{1}{c|}{val} & \multicolumn{1}{c|}{testA} & \multicolumn{1}{c|}{testB} & \multicolumn{1}{c|}{val} & \multicolumn{1}{c|}{testA} & \multicolumn{1}{c|}{testB} & \multicolumn{1}{c|}{val(u)} & \multicolumn{1}{c|}{test(u)} & \multicolumn{1}{c|}{val(g)} &  \\ 
\midrule
Traditional Full Fine-tuning &  &  &  &  &  &  &  &  &  &  \\ 
\midrule

$\text{ReSTR}_{[\text{CVPR 22}]}$~\cite{restr} & 67.2 & 69.3 & \multicolumn{1}{c|}{64.5} & 55.8 & 60.4 & \multicolumn{1}{c|}{48.3} & 54.5 & - & \multicolumn{1}{c|}{54.5} & 58.8 \\
$\text{CRIS}_{[\text{CVPR 22}]}$~\cite{cris} & 70.5 & 73.2 & \multicolumn{1}{c|}{66.1} & 62.3 & 68.1 & \multicolumn{1}{c|}{53.7} & 59.9 & 60.4 & \multicolumn{1}{c|}{-} & 63.8 \\
$\text{LAVT}_{[\text{CVPR 22}]}$~\cite{lavt} & 72.7 & 75.8 & \multicolumn{1}{c|}{68.8} & 62.1 & 68.4 & \multicolumn{1}{c|}{55.1} & - & - & \multicolumn{1}{c|}{60.5} & 64.9 \\
$\text{SEEM}_{[\text{NeurlPS 23}]}$~\cite{seem} & - & - & \multicolumn{1}{c|}{-} & - & - & \multicolumn{1}{c|}{-} & 65.6 & - & \multicolumn{1}{c|}{-} & - \\ 
$\text{VPD}_{[\text{ICCV 23}]}$~\cite{zhao2023unleashing} & 73.5 & - & \multicolumn{1}{c|}{-} & 63.9 & - & \multicolumn{1}{c|}{-} & 63.1 & - & \multicolumn{1}{c|}{-} & 66.8 \\ 
$\text{DMMI}_{[\text{ICCV 23}]}$~\cite{hu2023onetoonerethinkingreferringimage} & 74.1 & 77.1 & \multicolumn{1}{c|}{70.2} & 64.0 & 69.7 & \multicolumn{1}{c|}{57.0} & 63.5 & 64.2 & \multicolumn{1}{c|}{62.0} & 66.9 \\
$\text{ReLA}_{[\text{CVPR 23}]}$~\cite{liu2023gresgeneralizedreferringexpression} & 73.8 & 76.5 & \multicolumn{1}{c|}{70.2} & 66.0 & 71.0 & \multicolumn{1}{c|}{57.7} & 65.0 & 66.0 & \multicolumn{1}{c|}{62.7} & 67.7 \\
$\text{ReLA}_{[\text{CVPR 23}]}$~\cite{gres} & 73.8 & 76.5 & \multicolumn{1}{c|}{70.18} & 66.0 & 71.0 & \multicolumn{1}{c|}{57.7} & 65.0 & 66.0 & \multicolumn{1}{c|}{62.7} & 67.5 \\ 
$\text{CGFormer}_{[\text{CVPR 23}]}$~\cite{tang2023contrastive} & 74.8 & 77.3 & \multicolumn{1}{c|}{70.6} & 64.5 & 71.0 & \multicolumn{1}{c|}{57.1} & 64.7 & 65.1 & \multicolumn{1}{c|}{62.5} & 67.7\\ 
$\text{LISA-7B}_{[\text{CVPR 24}]}$~\cite{lai2023lisa} & 74.1 & 76.5 & \multicolumn{1}{c|}{71.1} & 62.4 & 67.4 & \multicolumn{1}{c|}{56.5} & 66.4 & {68.5} & \multicolumn{1}{c|}{-} & 67.9 \\ 
$\text{MagNet}_{[\text{CVPR 24}]}$~\cite{chng2023mask} & 75.2 & {78.2} & \multicolumn{1}{c|}{71.1} & 66.2 & 71.3 & \multicolumn{1}{c|}{58.1} & 65.4 & 66.2 & \multicolumn{1}{c|}{63.1} & 68.3 \\ 
$\text{ReMamber}_{[\text{ECCV 24}]}$~\cite{yang2024remamber} & 74.5 & 76.7 & \multicolumn{1}{c|}{70.9} & 65.0 & 70.8 & \multicolumn{1}{c|}{57.5} & 63.9 & 64.0 & \multicolumn{1}{c|}{-} & 67.9 \\ 

$\text{RISCLIP-B}_{[\text{NAACL 24}]}$~\cite{kim2024extendingclipsimagetextalignment} & 75.7 & 78.0 & \multicolumn{1}{c|}{72.5} & 69.2 & 73.5 & \multicolumn{1}{c|}{60.7} & 67.6 & 68.0 & \multicolumn{1}{c|}{-} & 70.6 \\

\midrule
\text{Parameter Efficient Tuning} &  &  & \multicolumn{1}{c|}{} &  &  & \multicolumn{1}{c|}{} &  &  & \multicolumn{1}{c|}{} &  \\ 
\midrule
$\text{ETRIS}_{ [\text{ICCV 23}]}$~\cite{xu2023bridging} & 70.5 & 73.5 & \multicolumn{1}{c|}{66.6} & 60.1 & 66.9 & \multicolumn{1}{c|}{50.2} & 59.8 & 59.9 & \multicolumn{1}{c|}{57.9} & 62.8 \\
$\text{BarLeRIa}_{[\text{ICLR 24}]}$~\cite{wang2023barleria} & 72.4 & 75.9  &  \multicolumn{1}{c|}{68.3} & 65.0 & 70.8 & \multicolumn{1}{c|}{56.9} & 63.4 & 63.8 & \multicolumn{1}{c|}{61.6} & 66.5 \\
$\text{DETRIS-B (Ours)}$ & {76.0} & {78.2} & \multicolumn{1}{c|}{{73.5}} & {68.9} & {74.0} & \multicolumn{1}{c|}{{61.5}} & {67.9} & {68.1} & \multicolumn{1}{c|}{{65.9}} & {70.4} \\

$\text{DETRIS-L (Ours)}$ & \textbf{77.3} & \textbf{79.0} & \multicolumn{1}{c|}{\textbf{75.2}} & \textbf{70.8} & \textbf{75.3} & \multicolumn{1}{c|}{\textbf{64.7}} & \textbf{69.3} & \textbf{70.2} & \multicolumn{1}{c|}{\textbf{67.9}} & \textbf{72.2} \\

\midrule
\text{With Mixed Training Data} &  &  & \multicolumn{1}{c|}{} &  &  & \multicolumn{1}{c|}{} &  &  & \multicolumn{1}{c|}{} &  \\ 
\midrule
$\text{PolyFormer-L\textsuperscript{*}}_{[\text{CVPR 23}]}$~\cite{polyformer} & 76.0 & 78.3 & \multicolumn{1}{c|}{73.3} & 69.3 & 74.6 & \multicolumn{1}{c|}{61.9}  & 69.2 & 70.2 & \multicolumn{1}{c|}{-} & 71.6 \\
$\text{UNINEXT-L\textsuperscript{*}}_{[\text{CVPR 23}]}$~\cite{uninext}& {80.3} & \textbf{82.6} & \multicolumn{1}{c|}{77.8} & 70.0 & 74.9 & \multicolumn{1}{c|}{62.6} & 73.4 & 73.7 & \multicolumn{1}{c|}{-} & 74.4 \\ 

$\text{DETRIS-L\textsuperscript{*} (Ours)}$ & \textbf{81.0} & {81.9} & \multicolumn{1}{c|}{\textbf{79.0}} & \textbf{75.2} & \textbf{78.6} & \multicolumn{1}{c|}{\textbf{70.2}} & \textbf{74.6} & \textbf{75.3} & \multicolumn{1}{c|}{{-}} & \textbf{77.2} \\

\bottomrule
\end{tabular}
\end{table*}

An up-sampling projector further transforms the multi-modal features \( F_{mm} \) and the sentence-level feature \( f_s \) to extract the cross-modal feature \( F_{c} \) and the transformed textual feature \( F_l \). \( f_s \) is first transformed into \( F_l \) through a linear transformation, then split and reshaped into weights and bias, enabling it to function as a Conv2D layer.
This Conv2D layer is used to transform the cross-modal representation into the final mask prediction.
The overall transformation is achieved using a 4$\times$ upsampling followed by convolution and linear projection:
\begin{equation}
\begin{aligned}
F_{c} &= \text{Conv}(\text{UpSample}(F_{mm})),\\
F_l &= \text{Linear}(f_s).
\end{aligned}
\end{equation}

\subsection{Training Objective}

Following CRIS~\cite{cris}, we adopt a text-to-visual contrastive loss, denoted as $\mathcal{L}_{\mathrm{con}}$, for the training objective of our model to optimize the alignment between text-derived features and their corresponding visual pixels.

This contrastive loss is designed to both enhance the connection between text features and corresponding visual pixels, and separate these text features from any unrelated visual elements. The text-to-pixel contrastive loss is mathematically articulated in the following manner:
\begin{equation}\begin{aligned}
&\mathcal{L}_{\mathrm{con}}^i\left(F_{c}^i,F_l\right)=\begin{cases}
-\log\left(\sigma\left(F_{c}^i\cdot F_l\right)\right), & i\in\mathcal{P} \\
-\log\left(1-\sigma\left(F_{c}^i\cdot F_l\right)\right), & i\in\mathcal{N}
\end{cases} \\
&\mathcal{L}_{\mathrm{con}}\left(F_{c},F_l\right)=\frac{1}{\left|\mathcal{P}\cup\mathcal{N}\right|}\sum_{i\in\mathcal{P}\cup\mathcal{N}}\mathcal{L}_{\mathrm{con}}^i \left(F_{c}^i,F_l\right),
\end{aligned}\end{equation}
where $\mathcal{P}$ and $\mathcal{N}$ denote the class of 1 and 0 in the ground truth, and $\sigma$ denotes the sigmoid function. The loss thus penalizes incorrect alignments between features and encourages the model to correctly match textual descriptions to their associated visual representations.

\begin{figure*}[ht]
    \centering
    \includegraphics[width=0.96\textwidth]{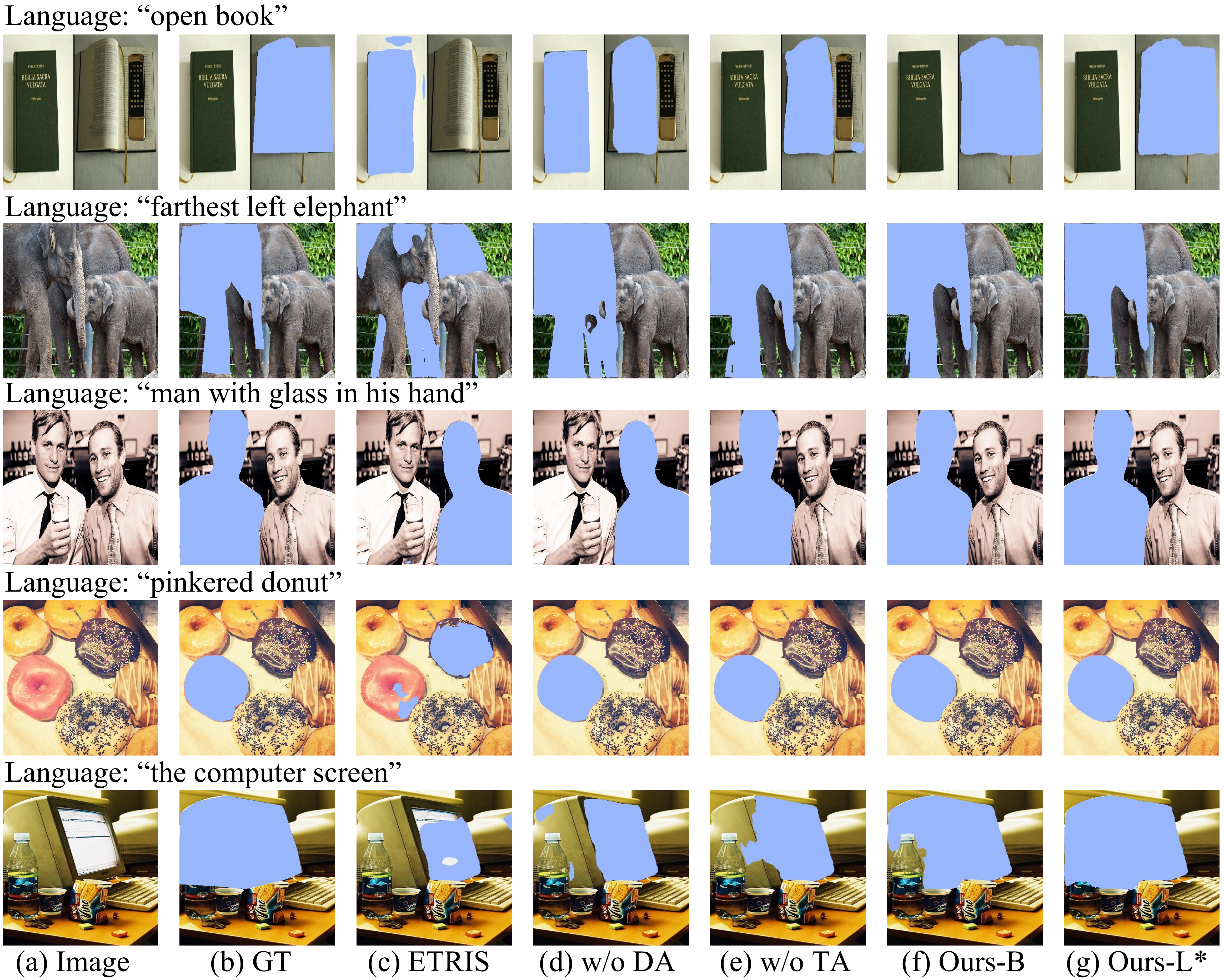}
    \caption{Qualitative results: (a) the input image; (b) the ground truth; (c) ETRIS; (d) DETRIS-B without Dense Aligner; (e) DETRIS-B without Text Adapter; (f) our proposed DETRIS-B; (g) DETRIS-L using mixed datasets. }
    \label{fig:visualization}
\end{figure*}

\section{Experiments}
\subsection{Datasets}
We employ three challenging referring image segmentation benchmarks in our experiments:
\begin{itemize}[leftmargin=*,noitemsep,nolistsep]
\item \textbf{RefCOCO~\cite{kazemzadeh2014referitgame}} is widely used as a benchmark for referring image segmentation. It comprises 19,994 images annotated with 142,210 referring expressions for 50,000 objects, which have been sourced from the MSCOCO dataset through a two-player game. The dataset is divided into four subsets, consisting of 120,624 training samples, 10,834 validation samples, 5,657 samples for test A, and 5,095 samples for test B, respectively. The average length of the expressions is 3.6 words, and each image contains a minimum of two objects.

\item \textbf{RefCOCO+~\cite{kazemzadeh2014referitgame}} dataset consists of 141,564 referring expressions associated with 49,856 objects in 19,992 images. The dataset is divided into four subsets: 120,624 train, 10,758 validation, 5,726 test A, and 4,889 test B samples. Notably, the RefCOCO+ dataset has been constructed to be more challenging than the RefCOCO dataset by excluding certain types of absolute location words. 

\item \textbf{G-Ref~\cite{yu2016modeling}} comprises 104,560 referring expressions associated with 54,822 objects in 26,711 images. The expressions in G-Ref were collected from Amazon Mechanical Turk and had an average length of 8.4 words, which included more words related to locations and appearances. We present results for both the Google and UMD partitioning methods for G-Ref.
\end{itemize}

\begin{table*}[htbp]
\caption{Comparison of Parameter-Efficient Tuning Methods Using DINO-B as Backbone on RefCOCO. To ensure fairness, we kept the original parameter settings from prior methods and also adjusted the size of rank to achieve comparable parameter counts. }
\label{tab:pet}
\centering
\begin{tabular}{l|ccc|c|r}
\toprule
\multirow{2}{*}{Method} & \multicolumn{3}{c|}{RefCOCO} & \multirow{2}{*}{Avg} & \multirow{2}{*}{Parameters (M)} \\ 
\cline{2-4} & \multicolumn{1}{c|}{val} & \multicolumn{1}{c|}{testA} & \multicolumn{1}{c|}{testB} & \\ 
\midrule

Full-Tuning & 65.1 & 68.1 & 61.4 & 64.9 & 149.97M \\
Fix Backbone & 74.9 & 77.1 & 72.0 & 74.7 & 0.00 M \\
Adapter \cite{houlsby2019parameter} & 71.2 & 73.3 & 68.3 & 70.9  & 1.98M \\

Compacter  \cite{karimi2021compacter} & 73.9 & 75.8 & 70.8 & 73.5  & 1.62M \\

LoRA \cite{hu2021lora} & 73.4 & 75.7 & 70.2 & 73.1 &  1.57M \\
ETRIS \cite{xu2023bridging} & 74.5 &  76.5 & 72.9 & 74.6  & 1.38M \\ \midrule
DETRIS-B (Ours) & 75.8 & 77.7 & 72.9 & 75.5  & 1.36M \\
DETRIS-B (Ours) (Default Setting) & 76.0 & 78.2 & 73.5 & 75.9  & 2.71M \\
\bottomrule
\end{tabular}		

\end{table*}
\subsection{Implementation Details}
In our experiments, DETRIS-B uses DINOv2-B/14 as the vision backbone, and DETRIS-L uses DINOv2-L/14. Both models employ the CLIP text encoder, with input images resized to 448x448 pixels. The Dense Aligner (dim=128) is applied at layers [1, 3, 5, 7, 9, 11] for DETRIS-B and [2, 6, 10, 14, 18, 22] for DETRIS-L. The Text Adapter (dim=64) is applied at layers [1, 3, 5, 7, 9, 11] in both models.
We train the framework end-to-end for 50 epochs using the Adam optimizer. The learning rate starts at 0.0001 and decays by 0.1 at epoch 35. DETRIS-B is trained on 2 A100 GPUs with a batch size of 32, while DETRIS-L uses 4 A100 GPUs with a batch size of 64 and an initial learning rate of 0.0002.
Performance is evaluated using mIoU, which measures the intersection-over-union between predicted masks and ground truth, following~\cite{wang2021cris}. 

\subsection{Main Results}
We compared our DETRIS models with previous RIS methods. As shown in Table~\ref{tab:main}, DETRIS-B achieves 70.4 IoU and DETRIS-L reaches 72.2, improving by 3.7\% and 6.6\% over the previous state-of-the-art. DETRIS-L outperforms all methods, especially on the challenging RefCOCO+ and G-Ref datasets.

In addition to comparing against full fine-tuning methods, we also evaluated our models in the context of parameter-efficient tuning methods. Table~\ref{tab:main} shows that DETRIS-B and DETRIS-L both outperform existing parameter-efficient tuning methods such as ETRIS and BarLeRIa. Specifically, DETRIS-B achieves a substantial improvement, and DETRIS-L further enhances performance, demonstrating the effectiveness of our method.

We also evaluated our models on mixed RefCOCO datasets to test generalization. As shown in Table~\ref{tab:main}, DETRIS-L achieves the highest average IoU of 77.2, outperforming methods like PolyFormer-L and UNINEXT-L. This highlights the robustness and effectiveness of our approach, particularly as data volume increases. Our DETRIS models offer significant improvements in both IoU and parameter efficiency over existing RIS methods.

\subsection{Qualitative Analysis}
As shown in Figure~\ref{fig:visualization}, we present qualitative results with different settings across various scenarios. Panel (c) shows the baseline ETRIS method, which struggles with accurate localization and segmentation. In contrast, panel (d) highlights DETRIS-B with only the Text Adapter, achieving improved segmentation for text-driven descriptions, while panel (e) demonstrates DETRIS-B with only the Dense Aligner, providing better visual-text alignment. Panel (f), combining both adapters, achieves the most accurate and balanced results, while panel (g) displays DETRIS-L trained on mixed datasets, offering the best overall performance.
Our methods significantly improve over ETRIS, especially in challenging scenarios, addressing limitations in object boundaries and ambiguous descriptions.

\subsection{Ablation Study}
\textbf{Comparison with Other Parameter-Efficient Tuning Methods.} 
We compare our Dense Aligner and Text Adapter (DETRIS-B) approach with other parameter-efficient tuning methods using DINO-B as the backbone. To ensure fairness, we retain the original parameter settings from prior methods and adjust rank size for comparable parameter counts.
As shown in Table~\ref{tab:pet}, DETRIS-B achieves superior performance while remaining efficient. Adapter reaches 70.9 with 1.98M parameters, Compacter 73.5 with 1.62M, LoRA 73.1 with 1.57M, and ETRIS 74.6 with 1.38M. Fully fine-tuning the misaligned DINO backbone degrades performance, underscoring the challenges of using DINO for RIS tasks.
Our DETRIS-B surpasses these methods, achieving 75.5 with 1.36M parameters (0.9\% of the backbone) by reducing Dense Aligners from 6 to 3. With 2.71M parameters (1.8\% of the backbone), DETRIS-B achieves the highest score of 75.9.
These results highlight the limitations of prior methods, which rely on pre-trained vision-text alignment and struggle in downstream RIS tasks with misaligned backbones. In contrast, our Dense Aligner actively learns cross-modal information during fine-tuning, overcoming these challenges effectively.

\begin{table}[htbp]
\caption{Ablation study on the components of DETRIS. DA stands for Dense Aligner, and TA denotes Text Adapter.}
\label{tab:abl}
\centering
\begin{tabular}{l|l|l|l|l|l}
\toprule
\multirow{2}{*}{DA} & \multirow{2}{*}{TA} & \multicolumn{3}{c|}{RefCOCO} &  \multirow{2}{*}{Avg} \\ 
\cline{3-5} & & \multicolumn{1}{c|}{val} & \multicolumn{1}{c|}{testA} & \multicolumn{1}{c|}{testB} & \\ 
\midrule
× & × & 74.9 & 77.1 & 71.9 & 74.6\\
\checkmark & × & 75.4 & 77.3 & 72.5 & 75.0  \\
× & \checkmark & 75.9 & 77.5 & 72.7 & 75.4   \\
\checkmark & \checkmark & 76.0 & 78.2 & 73.5 & 75.9   \\ 
\bottomrule
\end{tabular}
\end{table}

\textbf{Effect of Dense Aligner and Text Adapter.} We evaluated the Dense Aligner (DA) and Text Adapter (TA) through an ablation study on validation and test datasets. As shown in Table~\ref{tab:abl}, the baseline model without adapters achieves the lowest average score of 74.6. Adding the Dense Aligner slightly improves performance to 75.0, highlighting its role in enhancing visual feature extraction. The Text Adapter alone yields a greater improvement, raising the score to 75.4, demonstrating its importance in processing textual information. Combining both adapters achieves the highest score of 75.9, highlighting their combined effectiveness in improving cross-modal feature learning and segmentation.

\section{Conclusion}
In this work, we introduce DETRIS, a parameter-efficient tuning framework for referring image segmentation (RIS). Our approach adapts the pre-trained DINO model for RIS, focusing on aligning encoders not pre-trained for multi-modal tasks. We propose the Dense Aligner to enhance fine-grained visual-text alignment and use text adapters to augment the language encoder. Our method outperforms state-of-the-art fully fine-tuned models while being more scalable and efficient in parameter management.

\section*{Acknowledgments}
This work was supported by the Guangdong Provincial Natural Science Foundation Project under Grant 2022A1515011120.

\bibliography{aaai25}
\newpage
\appendix
\section{Appendix / Supplemental material}
\subsection{More comparative results}
\begin{figure}[htbp]
    \centering
    \includegraphics[width=1\linewidth]{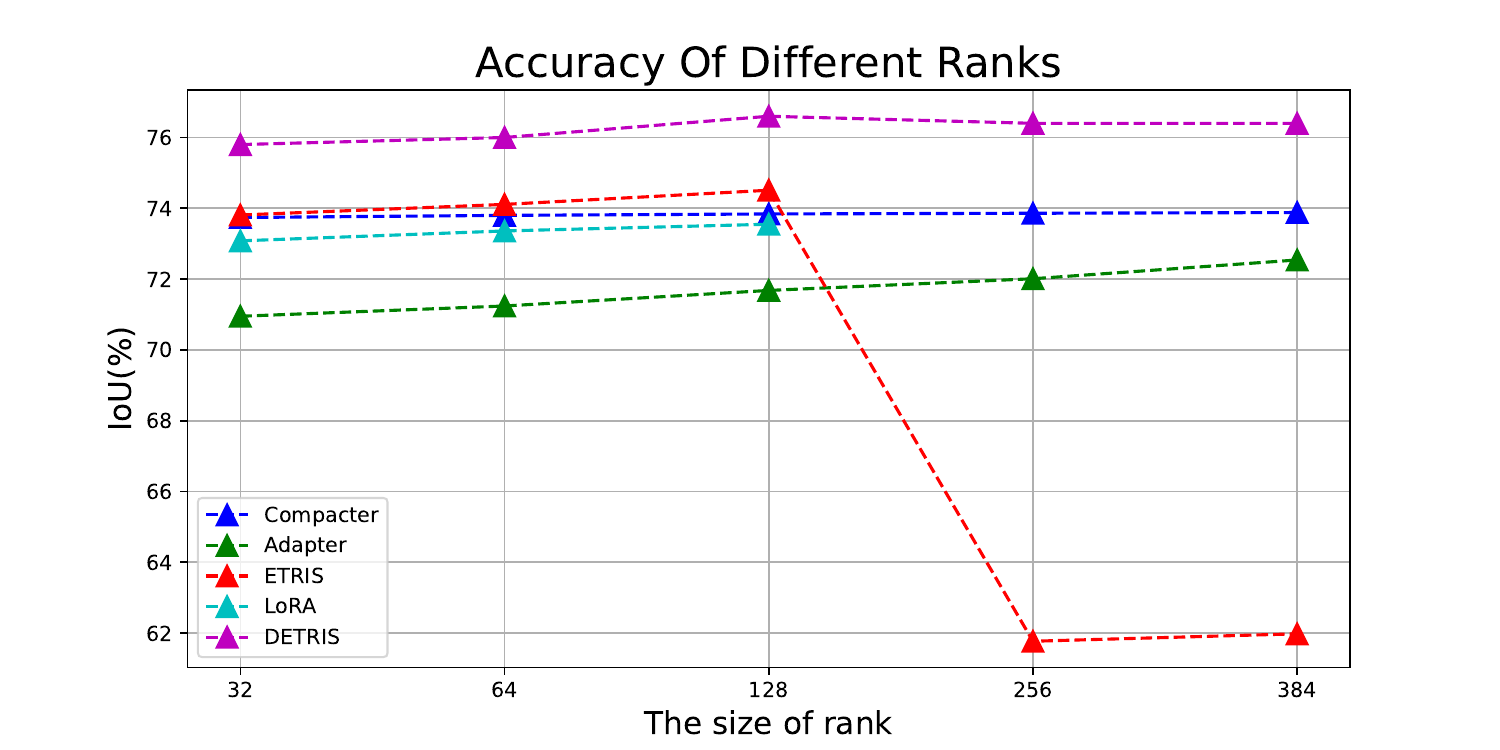}
    \caption{Ablation study of the Adapter’s rank and comparison with other Parameter-Efficient Tuning
Methods.}
    \label{fig:params}
\end{figure}

We investigated the effect of varying the dimensions of the adapters. The results of this study are illustrated in Figure~\ref{fig:params}. In this experiment, the adapters were fixed at the optimal layers identified in the previous experiment [1,3,5,7,9,11] while varying their dimensions. Configurations tested included combinations such as 64,64; 128,64; 128,128; and 256,128. The results, presented in Table~\ref{tab:dim_perf}, showed that the combination of 128,64, which corresponds to the chosen parameter setup for DETRIS-B, proved to be the most effective. Although increasing the dimensions to 128,128 or 256,128 introduced more parameters, the performance improvements were marginal and did not justify the additional computational cost.

\subsection{Qualitative Analysis}
\begin{figure*}[htbp]
    \centering
    \includegraphics[width=1\linewidth]{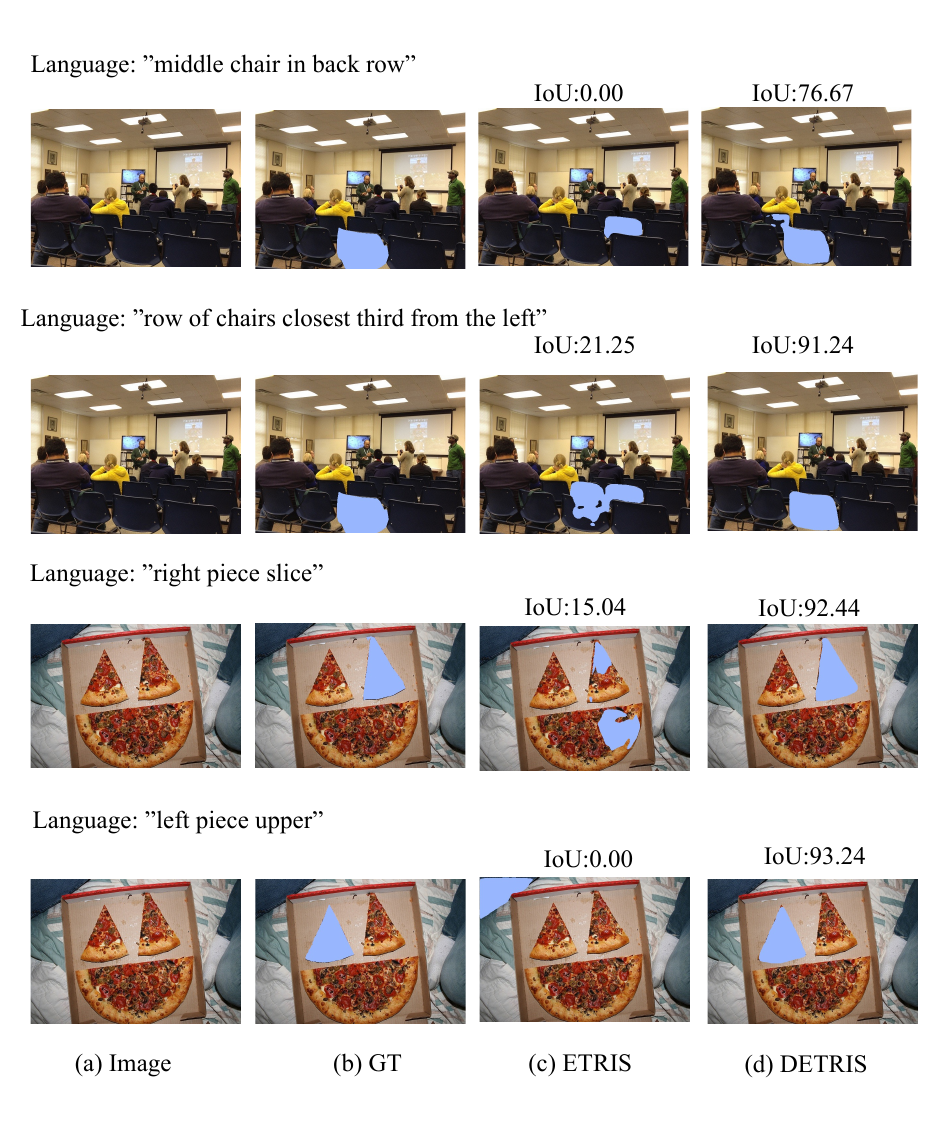}
    \caption{Comparison between DETRIS and state-of-the-art PET RIS method ETRIS.}
    \label{fig:visual_compare}
\end{figure*}

Figure~\ref{fig:visual_compare} illustrates a comparative analysis of qualitative results generated by DETRIS and ETRIS on a diverse set of test cases, highlighting DETRIS's superior performance in referring image segmentation tasks. Each subfigure presents the input image (a), ground truth (b), results from the baseline method ETRIS (c), and outputs from our DETRIS framework (d). The IoU scores, annotated for each result, quantitatively represent the model's performance. Notable examples include scenarios requiring precise identification of objects, such as "row of chairs closest third from the left," where DETRIS achieves an IoU of 76.67 compared to ETRIS's IoU of 15.04. Similarly, for "middle chair in back row," DETRIS demonstrates robust segmentation with an IoU of 91.24, significantly outperforming ETRIS's IoU of 21.25. These results emphasize the effectiveness of DETRIS's Dense Aligner and Text Adapter in enhancing cross-modal alignment and segmentation accuracy.

\begin{table}[!ht]
\caption{Ablation study of different dimensions (Dim) of Dense Aligner and Text Adapter.}
\label{tab:dim_perf}
\centering
\setlength{\tabcolsep}{1mm} 
\begin{tabular}{l|l|l|l|lll}
\toprule
\multirow{2}{*}{Visual Dim} & \multirow{2}{*}{Text Dim} & \multirow{2}{*}{Params}  & \multicolumn{3}{c}{RefCOCO} \\ \cline{4-6} 
&  & &  \multicolumn{1}{c|}{Val} &\multicolumn{1}{c|}{TestA} & \multicolumn{1}{c}{TestB} \\ \midrule
64 & 64 & 1.93M & 75.8 & \multicolumn{1}{l|}{77.9} & 73.3 \\ 
128 & 64 & 2.71M & 76.0 & \multicolumn{1}{l|}{78.2} & 73.5 \\ 
128 & 128 & 3.09M & 76.6 & \multicolumn{1}{l|}{78.2} & 73.3 \\ 
256 & 128 & 4.83M & 76.4 & \multicolumn{1}{l|}{78.4} & 73.5 \\ 
\bottomrule
\end{tabular}
\end{table}

\subsection{Limitations}

While our proposed DETRIS has outperformed fully fine-tuned models in terms of efficiency, scalability, and parameter management, our experiments have been limited to the RIS task. Future work should extend the validation to other multi-modal segmentation tasks to further confirm the versatility of our approach. Although DETRIS demonstrates strong performance by leveraging the Dense Aligner to effectively integrate the visual backbone network with textual global priors, our attempts to incorporate visual feature information into the text module have yielded mixed results. Specifically, we found that either the performance gains were minimal or the integration required a significant number of additional parameters, which goes against our goal of efficiency. Consequently, we have yet to identify an efficient and effective method for achieving robust visual-text integration within the text branch. Moreover, as multi-modal large-scale models advance, exploring open-vocabulary zero-shot referring image segmentation presents a promising avenue for research.
\end{document}